\documentclass[fleqn,10pt]{wlscirep}
\usepackage[utf8]{inputenc}
\usepackage[T1]{fontenc}
\usepackage{makecell}
\usepackage{tabularray}
\usepackage{multirow}
\usepackage{silence}
\title{DermaBench: A Clinician-Annotated Benchmark Dataset for Dermatology Visual Question Answering and Reasoning}

\author[1,*]{Abdurrahim Yilmaz}
\author[2,]{Ozan Erdem}
\author[3,]{Ece Gokyayla}
\author[4,]{Ayda Acar}
\author[5,]{Burc Bugra Dagtas}
\author[6,]{Dilara Ilhan Erdil}
\author[7,*]{Gulsum Gencoglan}
\author[1,*]{Burak Temelkuran}
\affil[1]{Imperial College London, Division of Systems Medicine, Department of Metabolism, Digestion, and Reproduction, United Kingdom}
\affil[2]{Istanbul Medeniyet University, Department of Dermatology and Venereology, Turkiye}
\affil[3]{Usak Research and Training Hospital, Department of Dermatology and Venereology, Turkiye}
\affil[4]{Ege University, Department of Dermatology and Venereology, Turkiye}
\affil[4]{Istanbul Research and Training Hospital, Department of Dermatology and Venereology, Turkiye}
\affil[5]{Ipswich Hospital, Department of Dermatology and Venereology, United Kingdom}
\affil[6]{Medicana Atakoy Hospital, Department of Dermatology and Venereology, Turkiye}

\affil[*]{Corresponding authors: Abdurrahim Yilmaz (a.yilmaz23@imperial.ac.uk), Gulsum Gencoglan (ggencoglan@gmail.com), Burak Temelkuran (b.temelkuran@imperial.ac.uk)}

\begin{abstract}
Vision--language models (VLMs) are increasingly important in medical applications; however, their evaluation in dermatology remains limited by datasets that focus primarily on image-level classification tasks such as lesion recognition. While valuable for recognition, such datasets cannot assess the full visual understanding, language grounding, and clinical reasoning capabilities of multimodal models. Visual question answering (VQA) benchmarks are required to evaluate how models interpret dermatological images, reason over fine-grained morphology, and generate clinically meaningful descriptions. We introduce DermaBench, a clinician-annotated dermatology VQA benchmark built on the Diverse Dermatology Images (DDI) dataset. DermaBench comprises 656 clinical images from 570 unique patients spanning Fitzpatrick skin types I--VI. Using a hierarchical annotation schema with 22 main questions (single-choice, multi-choice, and open-ended), expert dermatologists annotated each image for diagnosis, anatomic site, lesion morphology, distribution, surface features, color, and image quality, together with open-ended narrative descriptions and summaries, yielding approximately 14.474 VQA-style annotations. DermaBench is released as a metadata-only dataset to respect upstream licensing and is publicly available at Harvard Dataverse.
\end{abstract}

\begin{document}

\flushbottom
\maketitle

\thispagestyle{empty}

\section*{Background \& Summary}
Dermatology is one of the most visually driven specialties in medicine, relying heavily on pattern recognition across a wide spectrum of skin morphologies, colors, and anatomical sites. With the increasing adoption of teledermatology, mobile consultations, and electronic health records, vast quantities of dermatologic images are now captured daily in real-world settings. This surge of data offers an unprecedented opportunity for artificial intelligence (AI) to assist in diagnosis, triage, and education. However, despite these advances, the field lacks standardized, expert-annotated visual question answering (VQA) benchmark datasets that capture the true diversity of dermatologic conditions while maintaining clinically detailed labels and equitable representation of different skin tones.

Recent work in medical multimodal learning has produced a growing ecosystem of visual question answering (VQA) and vision–language datasets across medical fields such as radiology and pathology. Radiology datasets such as VQA-RAD comprise manually written questions but remain small (315 images, 3,515 QA pairs) and limited in scope, and their questions focus on modality or organ identification rather than clinical reasoning \cite{vqarad}. PathVQA expands scale but generates QA pairs automatically from textbook captions using rule-based NLP, producing shallow questions not aligned with real clinical tasks and no clinician involvement \cite{pathvqa}. More recent multilingual resources such as WoundCareVQA collect naturally occurring text–image pairs but target consumer health wound triage rather than medically curated disease classification, and lack structured dermatologic annotation or expert-defined ground-truth reasoning \cite{yim2025woundcarevqa}. Similarly, SLAKE provides semantic annotations and a medical knowledge graph, but focuses on radiology and CT/MRI imaging modalities rather than clinical dermatology .

In dermatology, most publicly available datasets have been developed for image-level classification tasks, such as lesion diagnosis or malignancy detection. Widely used datasets and related collections have enabled substantial progress in supervised image classification, particularly for melanoma recognition and differential diagnosis \cite{tschandl2018ham10000, combalia2024bcn20000, isicarchive, derm12345}. However, these datasets provide only a single diagnostic label per image and do not support evaluation of language grounding, fine-grained visual understanding, or clinical reasoning-capabilities that are central to emerging vision language models.

In contrast, dermatology-focused multimodal and visual question answering (VQA) datasets remain limited in trustworthiness, annotation depth, and clinical validity. DermatoLlama\cite{yilmaz2025resource} introduces a large dedicated test set (Eval1) comprising 210 dermoscopic images paired with 21,597 dermatologist-curated VQA pairs, providing a clinically grounded benchmark for assessing visual understanding and question answering performance in dermatology-focused VLMs. DermaVQA\cite{dermavqa} is a public dermatology VQA dataset, yet its answers are sourced from patient-consultation websites, resulting in variable accuracy and limited medical reliability, and its images are restricted to clinical photographs without expert verification or multimodal diversity. MM-Skin seeks to scale dermatology VQA by extracting captions from textbooks and using large language models to synthesize approximately 27k QA pairs, but its supervision is synthetically generated rather than clinician-authored, and blends automatically extracted captions with LLM-written conversations instead of dermatologist-verified ground truth \cite{mmskin}. More recent dermatology datasets with richer medical labels, such as SkinCon and SkinCAP (including SkinCoT and SkinCaRe), provide concept annotations, captions, or chain-of-thought reasoning; however, they do not offer an integrated VLM benchmark with standardized question taxonomy, multi-expert consensus labeling, or consistent evaluation of reasoning and fairness across skin tones and imaging modalities \cite{daneshjou2022skincon,zhou2024skincap}. Collectively, existing dermatology multimodal datasets exhibit three recurrent gaps: 
(1) absence of dermatologist-curated ground-truth VQA annotations for clinical dermatology images, 
(2) reliance on internet-scraped or synthetically generated supervision, and 
(3) lack of benchmarks explicitly designed to evaluate visual understanding, language grounding, reasoning, and fairness in dermatology-focused VLMs.

To address these limitations, we introduce DermaBench, a fully clinician-curated, multi-source dermatology benchmark designed expressly for evaluating the visual understanding and reasoning capabilities of vision–language models. Unlike prior datasets, DermaBench provides dermatologist-authored question–answer pairs in a structured VQA format, ensuring medically reliable ground truth for each item. The benchmark is built on Diverse Dermatology Image dataset \cite{ddistanford}, comprising 656 images from 570 patients, spanning clinical and dermoscopic modalities, each annotated by at least two expert dermatologists for various perspectives such as morphology, anatomical site. Importantly, every VQA item is written and validated by dermatologists, not synthesized by algorithms, overcoming the reliability limitations of DermaVQA and MM-Skin. As a result, DermaBench becomes the first dermatology dataset that combines multimodal imagery, expert-verified VQA annotations, standardized ontology-driven labels, and fairness-aware skin tone distributions, enabling rigorous evaluation of both perception and reasoning. This design makes DermaBench a high-trust benchmark for dermato-focused vision–language models, providing a foundation for reproducible, fair, and clinically aligned multimodal AI research. The dataset is publicly available at \url{https://doi.org/10.7910/DVN/Q4LBIW}.

\section*{Methods}
We selected the Diverse Dermatology Images (DDI) dataset as the primary component of DermaBench because it is the only publicly available dermatology image collection designed explicitly for skin tone diversity and fairness research. DDI contains clinical photographs spanning Fitzpatrick skin types I–VI, with each image labeled by dermatologists for both diagnosis and Fitzpatrick phototype. Its structure-80 diagnostic categories distributed across varied ethnicities, disease severities, and body sites-makes it a high-trust, clinically grounded resource. This diversity enables DermaBench to represent lesions from both light- and dark-skinned patients more equitably, ensuring that fairness considerations can be systematically evaluated in downstream vision and vision-language models. The DDI dataset was annotated by six dermatologists using the AnnotatorMed interface to create a dermatological VQA dataset (Figure 1), resulting in 14,474 image–question–answer pairs (Table 2).

%To broaden taxonomic diversity beyond the diagnostic spectrum represented in DDI, we incorporated 88 additional classes from Atlas Dermatologica, selecting one representative image per class. These images were re-annotated by our dermatology team using the same labeling taxonomy applied to DDI. Because Atlas Dermatologica does not include phototype metadata, Fitzpatrick labels were assigned using reference samples from the Fitzpatrick17k dataset to maintain consistent phototype mapping across all sources. The integration of these rare or underrepresented classes complements the broader but uneven class distribution in DDI, creating a more complete and balanced benchmark.

%Together, DDI contributes a clinically diverse and phototype-verified core, while Atlas Dermatologica adds breadth in diagnostic categories. This combined approach increases the representational completeness and fairness suitability of DermaBench.

\subsection*{Dataset Composition}
DermaBench was developed as a clinician-curated benchmark focused on diagnostic diversity, demographic fairness, and precise labeling. The dataset integrates two open-source repositories: (i) the Diverse Dermatology Images dataset, selected for its Fitzpatrick skin tone annotations and broad diagnostic classes
%, and (ii) Atlas Dermatologica, from which 88 representative classes were added and mapped to Fitzpatrick phototypes. 
In total, DermaBench contains 656 images from 570 unique patients, covering a range of skin tones, lesion types, and clinical contexts suitable for evaluating dermatologic visual reasoning.

Six expert dermatologists participated in the annotation process. Each image was labeled using a comprehensive, hierarchical question set designed to capture global image characteristics, dermatologic morphology, and clinically meaningful narrative descriptions. The full schema consists of 22 primary questions (Q0–Q21) (Table 1, Supplementary Material 2-3). 
Importantly, the question set itself was designed by three expert dermatologists, drawing directly from dermatologic examination frameworks and the morphology chapters of standard dermatology textbooks to ensure clinical fidelity and alignment with real-world diagnostic workflows. The development of this schema followed an iterative process: the full hierarchical question set was reviewed, tested, and revised across three separate design rounds, each time incorporating feedback from expert dermatologists to refine clarity, branching logic, and morphological completeness. Importantly, the question set itself was designed by three expert dermatologists, drawing directly from dermatologic examination frameworks and the morphology chapters of standard dermatology textbooks to ensure clinical fidelity and alignment with real-world diagnostic workflows \cite{plewig2022braun, james2016andrews, kang2019fitzpatrick, bolognia2024dermatology}. The development of this schema followed an iterative process: the full hierarchical question set was reviewed, tested, and revised across three separate design rounds, each time incorporating feedback from each of expert dermatologists to refine clarity, branching logic, and morphological completeness. To further ensure consistency in downstream annotations, the same three dermatologists prepared a consensus guideline outlining standardized interpretations for lesion morphology, distribution patterns, color terminology, and surface characteristics. This consensus document was used throughout the annotation phase to minimize inter-rater variability and maintain uniform application of the hierarchical schema across all annotators. These procedures collectively strengthen the reliability, reproducibility, and clinical interpretability of the DermaBench annotations.

\subsection*{Annotation Framework, AnnotatorMed}

All images were annotated using AnnotatorMed, an open-source medical data annotation interface developed within the SCALEMED framework \cite{yilmaz2025resource}. AnnotatorMed is a web-based, modular tool that renders dynamic question hierarchies through a context-aware interface. Sub-questions are automatically displayed only when their relevant parent options are selected, preventing screen clutter and reducing scrolling time. This user experience (UX) design allows physicians to complete each annotation efficiently without navigating unnecessary question branches, enabling rapid yet comprehensive labeling during clinical workflows.

\subsection*{Consensus and Quality Control}
Each image was annotated independently by two expert dermatologists using a predefined hierarchical question schema. Prior to annotation, three senior dermatologists developed a consensus checklist (Supplementary Material 1) that precisely defined the meaning and scope of each question and answer option, aligning terminology with standard dermatologic examination frameworks. This checklist was used throughout the annotation process to minimize inter-annotator variability. Annotation discrepancies were resolved during a structured consensus review phase, in which dermatologists jointly adjudicated differences in lesion morphology, distribution patterns, diagnostic interpretation, and Fitzpatrick skin tone assignment. Following consensus resolution, a final quality control step was performed by a dermatologist and an engineer, who jointly reviewed all annotations to identify residual dermatologic inconsistencies, reasoning errors, or structural issues in the annotation records. Images (n=10) with unresolved disagreements, insufficient visual quality, or ambiguous clinical content were excluded to ensure dataset reliability.

\subsection*{Annotation Efficiency and UX Performance}
AnnotatorMed’s dynamic rendering mechanism substantially improved annotation efficiency. By displaying sub-questions only when their parent options were selected, the interface reduced visual clutter and scrolling time by an estimated 50\%. Dermatologists completed a full annotation in an average of 3–5 minutes per image. The combination of structured single choice, multiple-choice and open-ended questions and narrative fields allowed simultaneous quantitative and descriptive data capture, supporting downstream use in classification, retrieval, and multimodal VQA evaluation.

All data follow a VQA-style JSON (Supplementary Material 2) schema that includes image path, modality tag, Fitzpatrick phototype, diagnostic and morphological labels, question category, answer text, annotator identifier, and additional metadata fields supporting fairness and reasoning analyses. The dataset is released for research use under a permissive license and is accompanied by documentation detailing the annotation protocol, consensus process, and recommended evaluation workflows. This unified structure enables seamless integration into multimodal training pipelines and provides a reproducible, clinically grounded benchmark for dermatology-focused vision and vision–language models.

\section*{Data Records}
DermaBench does not host or redistribute any dermatologic images. Instead, the dataset provides a structured metadata table on Harvard Dataverse (The dataset is publicly available at \url{https://doi.org/10.7910/DVN/Q4LBIW}) \cite{dermabench} containing all annotation fields and image identifiers required to reproduce experiments or link to the original image sources \cite{ddistanford}. This design respects the licensing constraints of the upstream datasets while enabling standardized benchmarking. It was released under a Creative Commons Attribution (CC BY-NC) license.

Researchers must obtain direct access to the images from the original repositories. For the Diverse Dermatology Images (DDI) dataset, image files and access procedures are available at \url{https://ddi-dataset.github.io/index.html#dataset}. The DDI dataset is released under a restricted academic-use license: (i) permission is granted to view and use the dataset without charge for personal, non-commercial research purposes only, and (ii) users may not distribute, publish, or reproduce any portion of the dataset without prior written permission from the Stanford School of Medicine. DermaBench therefore provides only metadata and does not copy or redistribute DDI images in any form.

%For Atlas Dermatologica images, users should refer to the respective public domain or institutional access pathways. Once images are obtained from their original sources, they can be linked to the DermaBench metadata via the provided identifiers (e.g., DDI\_ID or source filename) to reproduce all benchmarking experiments.

The metadata is distributed as a tabular CSV-style dataset, with one row per image and annotation fields corresponding to the hierarchical question schema. Because DermaBench uses a conditional annotation workflow, many sub-question fields remain intentionally empty when a parent question was not triggered. For instance, sub-questions related to the lesion location “scalp” appear only when the annotator selected “scalp” in Q12; otherwise those columns remain blank.

DermaBench does not include a predefined train–test split. Instead, a unified metadata file is provided to allow researchers to construct splits appropriate for classification, fairness auditing, VQA, or multimodal reasoning tasks.

\subsubsection*{Dataset Metadata}

The DermaBench metadata file includes the following fields:

\begin{itemize}
\item \textbf{DDI\_file\_ID:} Identifier linking to the corresponding DDI image.  
\item \textbf{questionX:} The text of each question in the annotation schema (Q0–Q21).  
\item \textbf{answerX:} The dermatologist’s answer to each question. Empty fields indicate questions not triggered by branching logic.  
\item \textbf{subquestion fields:} Hierarchical attributes (e.g., Q16.1, Q18.1). Present only when applicable.  
\end{itemize}

The metadata schema supports both high-level tasks (disease classification, fairness assessment, distribution analysis) and fine-grained multimodal tasks (visual question answering, long-form reasoning, semantic parsing). Its hierarchical organization preserves the full richness of dermatologic morphology while maintaining compatibility with VQA-style architectures.

\section*{Technical Validation}

Reliable evaluation of dermatology-focused AI systems requires benchmark datasets that are clinically trustworthy, demographically inclusive, and annotated with sufficient detail to support both diagnostic and reasoning-based tasks. The Diverse Dermatology Images (DDI) dataset was originally curated to meet this need-the first publicly accessible dermatology collection with expert-confirmed diagnoses and intentionally diverse Fitzpatrick skin types. DermaBench builds upon this foundation by adding structured, multi-level visual question answering (VQA) annotations, enabling assessment not only of diagnostic performance but also of multimodal reasoning and fairness across skin tones.

Clinically, AI has the potential to assist triage workflows by helping distinguish benign from malignant lesions, thereby reducing delays in dermatologic evaluation. However, dermatology datasets must represent all patient groups fairly. Because algorithmic biases often arise from underrepresentation, the DDI dataset's intentional inclusion of Fitzpatrick I--VI skin tones provides a validated substrate for fairness auditing, which DermaBench directly inherits and extends through its annotation schema.

\subsection*{Validation of Source Dataset (DDI)}

The images used in DermaBench that originate from DDI underwent rigorous clinical and pathology-based validation at the time of the original dataset curation. As described in the DDI dataset paper \cite{ddistanford}, all images were retrospectively selected by reviewing pathology-confirmed cases from Stanford Clinics between 2010--2020.  
A total of 656 images representing 570 unique patients were included. Each diagnosis was validated by both a expert dermatologist and a dermatopathologist using the corresponding biopsy report. Skin tone labels were assigned using in-person clinical assessments, demographic photographs, and review of the clinical images by two expert dermatologists. These procedures ensure high-fidelity diagnostic labels and accurate Fitzpatrick assignments, which form the basis for downstream fairness assessments in DermaBench.

\subsection*{Validation of Annotation Quality}

DermaBench extends the DDI dataset through structured annotation of dermatologic morphology, image quality, distribution patterns, color hierarchy, and higher-level VQA tasks. All annotations were performed independently by expert dermatologists using a hierarchical schema implemented in AnnotatorMed. The annotation workflow is developed to increase:

\begin{itemize}
    \item consistency through controlled vocabularies and predefined branching logic,
    \item precision through specialist-performed labeling,
    \item reproducibility through explicit documentation of all question and answer fields
\end{itemize}

Because the annotation interface displays sub-questions only when parent attributes are selected, missing entries in sub-columns reflect intentional branching rather than incomplete annotation.

\section*{Usage Notes}
The annotation framework is organized into three functional blocks, each corresponding to a distinct stage in dermatologic image interpretation:

\begin{itemize}
    \item \textbf{0. Not answered field (Q0)} \\
    Indicates whether the questions is answered or not.
    \item \textbf{1. Fundamental Image Questions (Q1--Q10)} \\
    These questions evaluate foundational attributes of each image, including:
    \begin{itemize}
        \item content type and modality,
        \item identifiability and framing,
        \item focus, illumination, and presence of artifacts,
        \item Fitzpatrick skin tone.
    \end{itemize}
    This block ensures accurate filtering of non-dermatologic images, supports image quality control, and enables fairness auditing across skin phototypes.

    \item \textbf{2. Detailed Dermatology VQA (Q11--Q20)} \\
    These questions encode structured, lesion-level dermatologic information using controlled vocabularies designed for both clinical interpretability and machine learning. They include:
    \begin{itemize}
        \item Q11: lesion count and multiplicity,
        \item Q12: anatomical location,
        \item Q13: distribution patterns (27 predefined options),
        \item Q14: lesion shape (21 morphologic categories),
        \item Q15: border characteristics (12 options),
        \item Q16: surface features (27 categories such as scale, crust, erosion, atrophy, xerosis),
        \item Q17: lesion size estimation,
        \item Q18: color hierarchy (11 chromatic subtypes),
        \item Q19--Q20: elementary lesion types (38 primary and secondary morphologies).
    \end{itemize}
    This block standardizes the descriptive lexicon used in dermatology and provides granular, machine-readable representations of complex visual features.

    \item \textbf{3. Summary Question (Q21)} \\
    Annotators provide a concise integrative description that:
    \begin{itemize}
        \item synthesizes all previously documented findings,
        \item highlights salient morphological patterns,
        \item contextualizes the lesion within plausible diagnostic considerations.
    \end{itemize}
    This summary supports downstream evaluation of model-generated clinical reporting.

\end{itemize}

Across all blocks, the schema balances categorical precision with descriptive flexibility. The hierarchical branching system ensures that annotators select only clinically relevant sub-options, reducing cognitive load and supporting consistent representation across annotators and image types. This structure enables DermaBench to support multimodal modeling tasks ranging from classification to long-form clinical reasoning.\\

DermaBench annotations were designed to be broadly compatible with the SkinCon\cite{daneshjou2022skincon} dermatologic concept taxonomy while remaining practical for VQA and LLM benchmarking. The concept labels are grounded in previously defined clinical terms\cite{bolognia2024dermatology}, but they are not intended to be exhaustive, nor do they assume perfectly clear-cut boundaries between concepts. In practice, dermatologic descriptors may exhibit differing annotations across datasets, including SkinCon, particularly because attributes. As a result of expanding the concept set from 48 to 197 concepts, partial differences in annotation granularity and alignment emerge between DermaBench and SkinCon, motivating the inclusion of explicit concept mappings. Core morphological attributes are aligned at the question level (Supplementary Table 1), including primary lesion morphology (Q19), secondary morphology (Q20), surface features (Q16), color (Q18), and distribution (Q13). Given the potential for variability in definitions and annotations, we recommend providing SkinCon annotations alongside corresponding DermaBench columns (Supplementary Table 1) when performing LLM benchmarking, enabling transparent comparison and more consistent cross-dataset analyses while preserving the VQA-oriented structure of DermaBench.

\paragraph{VQA Prompt for DermaBench}
This prompt is designed to benchmark vision--language models on DermaBench by standardizing how models interpret dermatological images and respond to structured clinical questions. It ensures consistent use of dermatological terminology and annotation rules by instructing the model to follow the definitions and guidelines provided in Supplementary Material 1. The prompt should be directly paired with the question definitions and answer options in Supplementary Material 2.\\

\textbf{Main Prompt:} \\
You are a board-certified dermatologist with expertise in clinical image interpretation. Carefully analyze the provided dermatological image and answer the question below using only the visual information in the image and standard dermatological knowledge.

Follow the instructions strictly based on the question type:

\begin{itemize}
    \item If \texttt{QUESTION\_TYPE} is \textbf{single-choice}, select \emph{one and only one} answer from the provided options and return the option text exactly as written.
    \item If \texttt{QUESTION\_TYPE} is \textbf{multi-choice}, select \emph{all} options that apply based on the image. Return the selected option texts as a comma-separated list, using the exact wording.
    \item If \texttt{QUESTION\_TYPE} is \textbf{open-ended}, write a concise, structured, and clinically meaningful report using standard dermatological terminology. Do not speculate or include information not supported by the image.
\end{itemize}

Do not assume patient history, symptoms, or metadata unless explicitly provided. If the image cannot be evaluated for the given question, use the dataset-defined response indicating that the image should not be considered.

\medskip
\noindent
\textbf{Question type:} \texttt{\{QUESTION\_TYPE\}} \\
\textbf{Question:} \texttt{\{QUESTION\_TEXT\}} \\
\textbf{Options:} \texttt{\{QUESTION\_OPTIONS\}}

\section*{Data availability}
The dataset is publicly available at \url{https://doi.org/10.7910/DVN/Q4LBIW}. The annotation data is published in XLSX format and the question set is published in JSON format.

\section*{Code availability}
There is no code or codebase in this study.
%Codes are available at https://github.com/abdurrahimyilmaz/derm12345

\bibliography{bibliograhpy}

\section*{Acknowledgements}
There is no acknowledgement. 

\section*{Funding}
Abdurrahim Yilmaz has been funded by the President’s PhD Scholarships of Imperial College London.

\section*{Author contributions statement}
Conceptualization, A.Y., O.E., E.G., and G.G.; methodology, A.Y., G.G., and B.T.; software, A.Y.; validation, G.G., and O.E.; formal analysis, A.Y., and B.T.; investigation, G.G. and O.E.; resources, G.G. and O.E.; data curation, A.Y., O.E., E.G. A.A., B.B.D., D.I.E., and  G.G.; writing-original draft preparation, A.Y. and B.T.; writing-review and editing, A.Y. and G.G.; visualization, A.Y. All authors have read and agreed to the published version of the manuscript.

\section*{Competing interests}
There is no competing interest.

\section*{Figures \& Tables}
\begin{figure}[htbp]
    \centering  \includegraphics[width=0.95\textwidth]{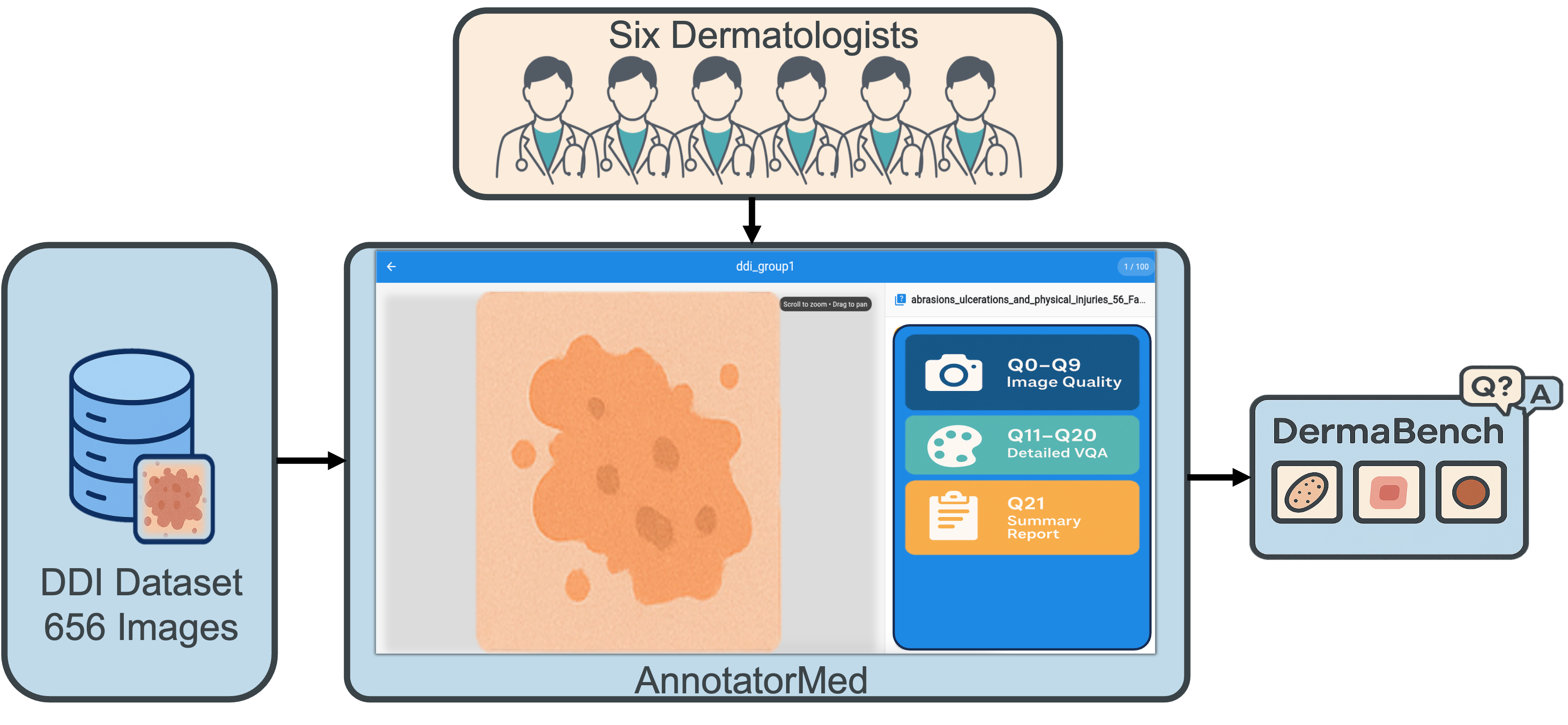}
  \caption{\textbf{Overview of the DermaBench framework.} 
  Data from DDI dataset % and \textbf{Atlas Dermatologica} 
  was annotated by six dermatologists using the AnnotatorMed interface \cite{yilmaz2025resource}, which dynamically renders hierarchical questions (Q0--Q21). 
  The resulting expert-validated annotations form the DermaBench dataset for standardized dermatology vision and VQA benchmarking.}
  \label{fig:framework}
\end{figure}

\begin{table}[htbp]
\centering
\small
\setlength{\tabcolsep}{5.2pt}
\begin{tabular}{@{}l l l c c c c c@{}}
\toprule
\textbf{Domain} & \textbf{Dataset} & \textbf{Modality} & \textbf{\#Images} & \textbf{\#QA Pairs} & \makecell{\textbf{Answer}\\\textbf{Length}}   \\
\midrule
Non-derm & VQA-RAD \cite{vqarad} & Radiology & 315 & 3.5K  & Open-ended \\
Non-derm & SLAKE \cite{slake} & Radiology & 642 & 14K  & Bilingual EN/ZH \\
Non-derm & PathVQA \cite{pathvqa} & Pathology & 5K & 32K  &  Multi-type Qs \\
\midrule
Derm & DermaVQA \cite{dermavqa} & Clinical (UGC) & 3.4K & 1.5K  & Multilingual, Reddit + IIYI \\
Derm & MM-Skin \cite{mmskin} & Various\textsuperscript{1}
 & $\sim$11K & 27K   & Adds textbook image--text pairs \\
Derm & Lightweight Derm VQA \cite{lightderm} & Clinical & 1,038 & $\sim$7.3K & 7 QA/img; low-resource focus \\
Derm & Test set of DermatoLlama \cite{yilmaz2025resource} & Dermoscopic & 210 & $\sim$21.5k & Dermoscopy only \\
\midrule
\textbf{Derm} & \textbf{DermaBench (Ours)} & Clinical  & \textbf{656} & \textbf{$\sim$14.4K} & 6 dermatologists; Fitzpatrick I--VI \\
\bottomrule
\end{tabular}
\caption{Comparison of DermaBench with representative medical VQA corpora. Various\textsuperscript{1}: Clinical, dermoscopic, pathology
}
\label{tab:comparison}
\end{table}

\begin{table}[h]
\centering
\caption{Distribution of question types and their functional role in the DermaBench annotation schema.}
\label{tab:questiontypes}
\begin{tabular}{lccc}
\hline
\textbf{Question Type} & \textbf{\#Questions}  & \textbf{Examples / Purpose} \\
\hline
Single-choice & 10 & Image category, illumination, focus, skin type \\
Multi-select & 14 & Lesion distribution, shape, border, surface, color, lesion types \\
Open-ended & 6 & Narrative quality notes, reports, explanations, summaries \\
\hline
\textbf{Total} & \textbf{30} &  \\
\hline
\end{tabular}
\end{table}

\end{document}